\documentclass[11pt]{article}
\usepackage{eacl2017}
\usepackage{times}
\usepackage{url}
\usepackage{latexsym}

\usepackage{times}
\usepackage{latexsym}
\usepackage{graphicx}
\usepackage{url}
\usepackage{amsmath}
\usepackage{verbatim}
\usepackage{tikz-qtree}
\usepackage{tikz}
\usetikzlibrary{shapes.geometric}
\usetikzlibrary{shapes.arrows}
\usetikzlibrary{patterns}
\usetikzlibrary{chains}
\usetikzlibrary{calc}

\eaclfinalcopy 
\def\eaclpaperid{***} 

\newcommand{\ignore}[1]{}

\title{Are Emojis Predictable? 
} 
  \author{Francesco Barbieri$^{\diamondsuit}$ ~~ Miguel Ballesteros$^{\spadesuit}$ ~~ Horacio Saggion$^{\diamondsuit}$\\
$^\diamondsuit$Large Scale Text Understanding Systems Lab, TALN Group\\ Universitat Pompeu Fabra, Barcelona, Spain \\
$^\spadesuit$IBM T.J Watson Research Center, U.S\\
{ \tt \{francesco.barbieri, horacio.saggion\}@upf.edu } \\
{ \tt miguel.ballesteros@ibm.com}
}

\date{}

\begin{document}
\maketitle
\begin{abstract}
Emojis are ideograms which are naturally combined with plain text to visually complement or condense the meaning of a message. Despite being widely used in social media, their underlying semantics have received little attention from a Natural Language Processing standpoint. 
In this paper, we investigate the relation between words and emojis, studying the novel task of predicting which emojis are evoked by text-based tweet messages.  We train several models based on Long Short-Term Memory networks (LSTMs) in this task.    
Our experimental results show that our neural model outperforms two  baselines as well as humans solving the same task, suggesting that computational models are able to better capture the underlying semantics of emojis.
\end{abstract}

\section{Introduction}

The advent of social media has brought along a novel way of communication where meaning is composed by combining short text messages and visual enhancements, the so-called \textit{emojis}. This visual language
is as of now a {\it de-facto} standard for online communication, available not only in Twitter, but also in other large online platforms such as Facebook, Whatsapp, or Instagram. 

Despite its status as language form, emojis have been so far scarcely studied from a Natural Language Processing (NLP) standpoint. 
Notable exceptions include studies focused on emojis' semantics and usage \cite{aoki2011method,barbieri2016revealing,barbieri2016cosmopolitan,emoji2016LREC,eisneremoji2vec2016,ljubevsic2016global}, or sentiment \cite{novak2015sentiment}. However, the interplay between text-based messages and emojis remains
virtually unexplored. This paper aims to fill this gap by investigating the relation between words and emojis, studying the problem of predicting which emojis are evoked by text-based tweet messages.

\newcite{miller2016blissfully} performed an evaluation asking human annotators the meaning of emojis, and the sentiment they evoke. People do not always have the same understanding of emojis, indeed, there seems to exist multiple interpretations of their meaning beyond their designer's intent or the physical object they evoke\footnote{https://www.washingtonpost.com/news/the-intersect/wp/2016/02/19/the-secret-meanings-of-emoji/}. Their main conclusion was that emojis can lead to misunderstandings.
The ambiguity of emojis raises an interesting question in human-computer interaction: how can we teach an artificial agent to correctly interpret and recognise emojis' use in spontaneous conversation?\footnote{http://www.dailydot.com/debug/emoji-miscommunicate/} The main motivation of our research is that an artificial intelligence system that is able to predict emojis could contribute to better natural language understanding \cite{novak2015sentiment} and thus to different natural language processing tasks such as generating emoji-enriched social media content, enhance emotion/sentiment analysis systems, and improve retrieval of social network material. 

In this work, we employ a state of the art classification framework to automatically predict the most likely emoji a Twitter message evokes. The model is based on Bidirectional Long Short-term Memory Networks (BLSTMs) with both standard lookup word representations and character-based representation of tokens. We will show that the BLSTMs outperform a bag of words baseline, a baseline based on semantic vectors,  and human annotators in this task.

\section{Dataset and Task}
\label{sec:dataset}

\textbf{Dataset:}
We retrieved 40 million tweets with the Twitter APIs\footnote{https://dev.twitter.com}. Tweets were posted between October 2015 and May 2016 geo-localized in the United States of America.
We removed all hyperlinks from each tweet, and lowercased all textual content in order to reduce noise and sparsity.
\begin{table}
\centering
\setlength{\tabcolsep}{2pt} 
\renewcommand{\arraystretch}{1} 
\scalebox{0.90}{
\begin{tabular}{|cccccccccc|} \hline
\includegraphics[height=0.42cm,width=0.42cm]{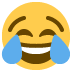} & \includegraphics[height=0.42cm,width=0.42cm]{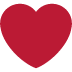} & \includegraphics[height=0.42cm,width=0.42cm]{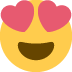} & \includegraphics[height=0.42cm,width=0.42cm]{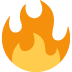} & \includegraphics[height=0.42cm,width=0.42cm]{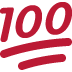} & \includegraphics[height=0.42cm,width=0.42cm]{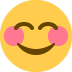} & \includegraphics[height=0.42cm,width=0.42cm]{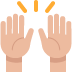} & \includegraphics[height=0.42cm,width=0.42cm]{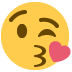} & \includegraphics[height=0.42cm,width=0.42cm]{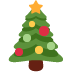} & \includegraphics[height=0.42cm,width=0.42cm]{1f495} \\ 
100.7 & 89.9 & 59 & 33.8 & 28.6 & 27.9 & 22.5 & 21.5 & 21 & 20.8 \\ \hline
\includegraphics[height=0.42cm,width=0.42cm]{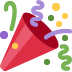} & \includegraphics[height=0.42cm,width=0.42cm]{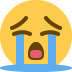} & \includegraphics[height=0.42cm,width=0.42cm]{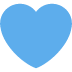} & \includegraphics[height=0.42cm,width=0.42cm]{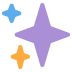} & \includegraphics[height=0.42cm,width=0.42cm]{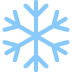} & \includegraphics[height=0.42cm,width=0.42cm]{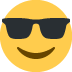} & \includegraphics[height=0.42cm,width=0.42cm]{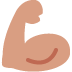} & \includegraphics[height=0.42cm,width=0.42cm]{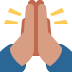} & \includegraphics[height=0.42cm,width=0.42cm]{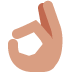} & \includegraphics[height=0.42cm,width=0.42cm]{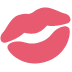} \\ 
19.5 & 18.6 & 18.5 & 17.5 & 17 & 16.1 & 15.9 & 15.2 & 14.2 & 10.9 \\
\hline\end{tabular}
}
\caption{\label{tab:freq} The 20 most frequent emojis that we use in our experiments and the number of thousand tweets they appear in.}
\end{table}
From the dataset, we selected tweets which include {\em one and only one} of the 20 most frequent emojis, resulting in a final dataset\footnote{Available at http://sempub.taln.upf.edu/tw/eacl17} composed of 584,600 tweets. In the experiments we also consider the subsets of the 10 (502,700 tweets) and 5 most frequent emojis (341,500 tweets). See Table~\ref{tab:freq} for the 20 most frequent emojis that we consider in this work.

\noindent\textbf{Task}: We remove the emoji from the sequence of tokens and use it as a label both for training and testing. The task for our machine learning models is to predict the single emoji that appears in the input tweet. 

\section{Models}
\label{sec:models}
In this Section, we present and motivate the models that we use to predict an emoji given a tweet. The first model is an architecture based on Recurrent Neural Networks (Section \ref{sec:model}) and the second and third are the two baselines (Section \ref{bow} and \ref{avg}). The two major differences between the RNNs and the baselines, is that the RNNs take into account sequences of words and thus, the entire context.

\subsection{Bi-Directional LSTMs}
\label{sec:model}
Given the proven effectiveness and the impact of recurrent neural networks in different tasks \cite[inter-alia]{DBLP:journals/corr/ChungGCB14,vinyals:2015,BahdanauCB14,lstmacl15,lample-EtAl:2016:N16-1,wang-EtAl:2016:N16-12}, which also includes modeling of tweets \cite{tweet2vec}, our emoji prediction model is based on bi-directional Long Short-term Memory Networks \cite{hochreiter:1997,graves:2005}.
The B-LSTM can be formalized as follows:
\begin{align*}
\mathbf{s} = \max \left\{\mathbf{0}, \mathbf{W}[\mathbf{fw}; \mathbf{bw}] + \mathbf{d}\right\}
\end{align*}
where $\mathbf{W}$ is a learned parameter matrix, $\mathbf{fw}$ is the forward LSTM encoding of the message, $\mathbf{bw}$ is the backward LSTM encoding of the message, and $\mathbf{d}$ is a bias term, then passed through a component-wise ReLU. The vector $\mathbf{s}$ is then used to compute the probability distribution of the emojis given the message as:
\begin{align*}
p(e \mid \mathbf{s}) = \frac{\exp \left( \mathbf{g}_{e}^{\top} \mathbf{s} + q_{e} \right)}{\sum_{e' \in \mathcal{E}} \exp \left( \mathbf{g}_{e'}^{\top} \mathbf{s} + q_{e'} \right)}
\end{align*}
where $\mathbf{g}_{e'}$ is a column vector representing the (output) embedding\footnote{The output embeddings of the emojis have 100 dimensions.} of the emoji $e$, and $q_e$ is a bias term for the emoji $e$. The set $\mathcal{E}$ represents the list of emojis. 
The loss/objective function the network aims to minimize is the following:
\begin{align*}
Loss = -log (p(e_m \mid \mathbf{s}))
\end{align*}

where $\mathit{m}$ is a tweet of the training set $\mathcal{T}$, $\mathbf{s}$ is the encoded vector representation of the tweet and $\mathit{e_m}$ is the emoji contained in the tweet $\mathit{m}$. 
The inputs of the LSTMs are word embeddings\footnote{100 dimensions.}. Following, we present two alternatives explored in the experiments presented in this paper. 
 
\noindent\textbf{Word Representations}: We generate word embeddings which are learned together with the updates to the model. We stochastically replace (with \emph{p} = 0.5) each word that occurs only once in the training data with a fixed represenation (out-of-vocabulary words vector). When we use pretrained word embeddings, these are concatenated with the learned vector representations obtaining a final representation for each word type. This is similar to the treatment of word embeddings by \newcite{lstmacl15}. 

\noindent\textbf{Character-based Representations}: We compute character-based continuous-space vector embeddings \cite{ling:2015,ballesteros-dyer-smith:2015:EMNLP} of the tokens in each tweet using, again, bidirectional LSTMs. The character-based approach learns  representations for words that are orthographically similar,  thus, they should be able  to handle different alternatives of the same word type occurring in social media. 

\subsection{Baselines}
In this Section we describe the two baselines. Unlike the previous model, the baselines do not take into account the word order. However, in the second baseline (Section \ref{avg}) we abstract on the plain word representation using semantic vectors, previously trained on Twitter data.

\subsubsection{Bag of Words}
\label{bow}
We applied a bag of words classifier as baseline, since it has been successfully employed in se\-veral classification tasks, like sentiment analysis and topic modeling \cite{wallach2006topic,blei2012probabilistic,titov2008modeling,maas2011learning,davidov2010semi}.
We represent each message with a vector of the most informative tokens (punctuation marks included) selected using term frequency$-$inverse document frequency (TF-IDF). We employ a L2-regularized logistic regression classifier to make the predictions.

\subsubsection{Skip-Gram Vector Average}
\label{avg}
We train a Skip-gram model \cite{mikolov2013exploiting} learned from 65M Tweets (where testing instances have been removed) to learn Twitter semantic vectors. Then, we build a model (henceforth, AVG) which represents each message as the average of the vectors corresponding to each token of the tweet. 
Formally, each message $\mathit{m}$ is represented with the vector $\mathit{V_{m}}$:
\begin{align*}
Vm = \frac{\sum_{t \in T_{m}} S_{t}}{|T_{m}|}
\end{align*}
\noindent Where $\mathit{T_{m}}$ are the set of tokens included in the message $\mathit{m}$, $\mathit{S_{t}}$ is the vector of token $\mathit{t}$ in the Skipgram model, and $\mathit{|T_{m}|}$ is the number of tokens in $\mathit{m}$.
After obtaining a representation of each message, we train a L2-regularized logistic regression, (with $\varepsilon$ equal to 0.001).

\section{Experiments and Evaluation}
\label{sec:exp}
In order to study the relation between words and emojis, we performed two different experiments. In the first experiment, we compare our machine learning models, and in the second experiment, we pick the best performing system and compare it against humans.

\subsection{First Experiment}
This experiment is a classification task, where in each tweet the unique emoji is removed and used as a label for the entire tweet. We use three datasets, each containing the 5, 10 and 20 most frequent emojis (see Section \ref{sec:dataset}). 
We analyze the performance of the five models described in Section \ref{sec:models}: a bag of words model, a Bidirectional LSTM model with character-based representations (char-BLSTM), a Bidirectional LSTM model with standard lookup word representations (word-BLSTM). The latter two were trained with/without pretrained word vectors. 
To pretrain the word vectors, we use a modified skip-gram model \cite{Ling:2015:naacl} trained on the English Gigaword corpus\footnote{https://catalog.ldc.upenn.edu/LDC2003T05} version 5.

We divide each dataset in three parts, training (80\%), development (10\%) and testing (10\%). The three subsets are selected in sequence starting from the oldest tweets and from the training set since automatic systems are usually trained on past tweets, and need to be robust to future topic variations. 

\begin{table}
\centering
\setlength{\tabcolsep}{4pt} 
\renewcommand{\arraystretch}{1} 
\scalebox{0.9}{
\begin{tabular}{|r|ccc|ccc|ccc|}
\hline 
 & \multicolumn{3}{c|}{\textbf{5}} 
 & \multicolumn{3}{c|}{\textbf{10}}
 & \multicolumn{3}{c|}{\textbf{20}} 
 \tabularnewline
 & \textbf{P} & \textbf{R} & \textbf{F1}  & \textbf{P} & \textbf{R} & \textbf{F1} &  \textbf{P} & \textbf{R} & \textbf{F1}  \tabularnewline
\hline 
\textbf{BOW}  & .59 & .60 & .58  & .43 & .46 & .41 & .32 & .34 & .29 \\
\textbf{AVG}  & .60 & .60 & .57  & .44 & .47 & .40 & .34 & .36 & .29 \\
\hline
\textbf{W} & .59 & .59 & .59  & .46 & .46 & .46 & .35 & .36 & .33  \\
\textbf{C} & .61 & .61 & .61  & .44 & .44 & .44 & .36 & .37 & .32  \\
\textbf{W+P} & .61 & .61 & .61 & .45 & .45 & .45 & .34 & .36 & .32 \\
\textbf{C+P} & \textbf{.63} & \textbf{.63} & \textbf{.63}  & \textbf{.48} & \textbf{.47} & \textbf{.47} & \textbf{.42} & \textbf{.39} & \textbf{.34}  \\

\hline 
\end{tabular}
}
\caption{\label{tab:results} Results of 5, 10 and 20 emojis. Precision, Recall, F-measure. BOW is bag of words, AVG is the Skipgram Average model, C refers to char-BLSTM and W refers to word-BLSTM. +P refers to pretrained embeddings.}
\end{table}

Table~\ref{tab:results} reports the results of the five models and the 
baseline. All  neural models outperform the baselines in all the experimental setups. However, the BOW and AVG are quite competitive, suggesting that most emojis come along with specific words (like the word {\em love} and the emoji \includegraphics[height=0.32cm,width=0.32cm]{2764}).
However, considering sequences of words in the models seems important for encoding the meaning of the tweet and therefore contextualize the emojis used. Indeed, the B-LSTMs models always outperform BOW and AVG. The character-based model with pretrained vectors is the most accurate at predicting emojis. 
The character-based model seems to capture orthographic variants of the same word in social media.
Similarly, pretrained vectors allow to initialize the system with unsupervised pre-trained semantic knowledge \cite{Ling:2015:naacl}, which helps to achieve better results.

\begin{table}[!ht]
\centering
\scalebox{0.95}{
\begin{tabular}{|c|ccc|c|c|} \hline
\textbf{Emoji} & \textbf{P} & \textbf{R} & \textbf{F1} & \textbf{Rank} & \textbf{Num}  \\ \hline
\includegraphics[height=0.42cm,width=0.42cm]{1f602} & 0.48 & \textbf{0.74} & \textbf{0.58} & 2.12 & 783 \\ 
\includegraphics[height=0.42cm,width=0.42cm]{2764} & 0.32 & \textbf{0.74} & 0.45 & \textbf{1.59} & 757 \\ 
\includegraphics[height=0.42cm,width=0.42cm]{1f60d} & 0.35 & 0.22 & 0.27 & 3.58 & 470 \\ 
\includegraphics[height=0.42cm,width=0.42cm]{1f60a} & 0.31 & 0.15 & 0.21 & 4.2 & 260 \\ 
\includegraphics[height=0.42cm,width=0.42cm]{1f60e} & 0.24 & 0.1 & 0.14 & 4.39 & 212 \\ 
\includegraphics[height=0.42cm,width=0.42cm]{1f525} & 0.46 & 0.49 & 0.47 & 3.76 & 207 \\ 
\includegraphics[height=0.42cm,width=0.42cm]{1f495} & 1 & 0 & 0.01 & 4.69 & 206 \\ 
\includegraphics[height=0.42cm,width=0.42cm]{1f4af} & 0.44 & 0.19 & 0.27 & 5.15 & 200 \\ 
\includegraphics[height=0.42cm,width=0.42cm]{1f4aa} & 0.44 & 0.54 & 0.48 & 4.71 & 165 \\ 
\includegraphics[height=0.42cm,width=0.42cm]{1f64c} & 0.33 & 0.11 & 0.17 & 5.79 & 150 \\ 
\includegraphics[height=0.42cm,width=0.42cm]{1f618} & 0.3 & 0.12 & 0.17 & 5.78 & 148 \\ 
\includegraphics[height=0.42cm,width=0.42cm]{1f499} & 0.54 & 0.11 & 0.18 & 6.73 & 131 \\ 
\includegraphics[height=0.42cm,width=0.42cm]{2728} & 0.45 & 0.19 & 0.27 & 6.43 & 120 \\ 
\includegraphics[height=0.42cm,width=0.42cm]{1f48b} & \textbf{0.56} & 0.09 & 0.15 & 7.58 & 112 \\ 
\includegraphics[height=0.42cm,width=0.42cm]{1f44c} & 0.2 & 0.01 & 0.02 & 9.01 & 110 \\ 
\includegraphics[height=0.42cm,width=0.42cm]{1f64f} & 0.46 & 0.33 & 0.39 & 5.83 & 108 \\ 
\includegraphics[height=0.42cm,width=0.42cm]{1f62d} & 0.5 & 0.08 & 0.13 & 4.9 & 105 \\ 
\includegraphics[height=0.42cm,width=0.42cm]{1f389} & 0.32 & 0.25 & 0.28 & 6.13 & 89 \\ 
\includegraphics[height=0.42cm,width=0.42cm]{2744} & 0.44 & 0.53 & 0.48 & 5.35 & 34 \\ 
\includegraphics[height=0.42cm,width=0.42cm]{1f384} & 0.22 & 0.67 & 0.33 & 1.67 & 3 \\ \hline
\end{tabular}
}
\caption{\label{tab:best}Precision, Recall, F-measure, Ranking and occurrences in the test set of the 20 most frequent emojis using char-BLSTM + Pre.}
\end{table}


\paragraph{Qualitative Analysis of Best System:}
\label{sec:exp2}
We analyze the performances of the char-BLSTM with pretrained vectors on the 20-emojis dataset, as it resulted to be the best system in the experiment presented above. In Table~\ref{tab:best} we report Precision, Recall, F-measure and Ranking\footnote{The Ranking is a number between 1 and 20 that represents the average number of emojis with higher probability than the gold emoji in the probability distribution of the classifier.} of each emoji. We also added in the last column the occurrences of each emoji in the test set. 

The frequency seems to be very relevant. The Ranking of the most frequent emojis is lower than the Ranking of the rare emojis. This means that if an emoji is frequent, it is more likely to be on top of the possible choices even if it is a mistake. On the other hand, the F-measure does not seem to depend on frequency, as the highest F-measures are scored by a mix of common and uncommon emojis (\includegraphics[height=0.32cm,width=0.32cm]{1f602.png},
\includegraphics[height=0.32cm,width=0.32cm]{2764.png},
\includegraphics[height=0.32cm,width=0.32cm]{1f525.png}, and
\includegraphics[height=0.32cm,width=0.32cm]{2744.png}) which are respectively the first, second, the sixth and the second last emoji in terms of frequencies.

The frequency of an emoji is not the only important variable to detect the emojis properly; it is also important whether in the set of emojis there are emojis with similar semantics. If this is the case the model prefers to predict the most frequent emojis. This is the case of the \includegraphics[height=0.32cm,width=0.32cm]{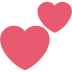} emoji that is almost never predicted, even if the Ranking is not too high (4.69). The model prefers similar but most frequent emojis, like \includegraphics[height=0.32cm,width=0.32cm]{2764.png} (instead of \includegraphics[height=0.32cm,width=0.32cm]{1f495.png}). The same behavior is observed for the \includegraphics[height=0.32cm,width=0.32cm]{1f499.png} emoji, but in this case the performance is a bit better due to some specific words used along with the blue heart: ``blue'', ``sea'' and words related to childhood (e.g. ``little'' or ``Disney'').

Another interesting case is the Christmas tree emoji \includegraphics[height=0.32cm,width=0.32cm]{1f384}, that is present only three times in the test set (as the test set includes most recent tweets and Christmas was already over; this emoji is commonly used in tweets about Christmas). The model is able to recognize it twice, but missing it once. The correctly predicted cases include the word ``Christmas''; and it fails to predict: \emph{``getting into the holiday spirit with this gorgeous pair of leggings today ! \#festiveleggings''}, since there are no obvious clues (the model chooses \includegraphics[height=0.32cm,width=0.32cm]{2764.png} instead probably because of the intended meaning of ``holiday'' and ``gorgeous''.).

In general the model tends to confuse similar emojis to  \includegraphics[height=0.32cm,width=0.32cm]{2764.png} and
\includegraphics[height=0.32cm,width=0.32cm]{1f602.png}, probably for their higher frequency and also because they are used in multiple contexts. An interesting phenomenon is that \includegraphics[height=0.32cm,width=0.32cm]{1f62d.png} is often confused with \includegraphics[height=0.32cm,width=0.32cm]{1f602.png}. The first one represent a small face crying, and the second one a small face laughing, but the results suggest that they appear in similar tweets. The punctuation and tone used is often similar (many exclamation marks and words like \emph{``omg''} and \emph{``hahaha''}).
Irony may also play a role to explain the confusion, e.g. \emph{``I studied journalism and communications , I'll be an awesome speller! Wrong.  \includegraphics[height=0.32cm,width=0.32cm]{1f62d.png} haha so much fun''.}

\begin{table}
\centering
\scalebox{0.95}{
\begin{tabular}{|c|c|c|c|c|c|c|c|c|}
\cline{2-7} 
\multicolumn{1}{c|}{} & \multicolumn{3}{c|}{\textbf{Humans}} & \multicolumn{3}{c|}{\textbf{B-LSTM}} \tabularnewline
\hline 
\textbf{Emo} & \textbf{P} & \textbf{R} & \textbf{F1} & \textbf{P} & \textbf{R} & \textbf{F1} \tabularnewline
\hline 
\includegraphics[height=0.42cm,width=0.42cm]{1f602.png} & 0.73 & 0.56 & 0.63 & 0.7 & 0.84 & \textbf{0.77}  \\
\includegraphics[height=0.42cm,width=0.42cm]{2764.png} & 0.53 & 0.51 & 0.52 & 0.61 & 0.78 & \textbf{0.69}  \\
\includegraphics[height=0.42cm,width=0.42cm]{1f60d.png} & 0.43 & 0.38 & \textbf{0.4} & 0.52 & 0.3 & 0.38 \\
\includegraphics[height=0.42cm,width=0.42cm]{1f4af.png} & 0.19 & 0.4 & 0.26 & 0.62 & 0.26 & \textbf{0.37}  \\
\includegraphics[height=0.42cm,width=0.42cm]{1f525.png} & 0.24 & 0.26 & 0.25 & 0.66 & 0.51 & \textbf{0.58}  \\ \hline \hline
\textbf{Avg} & 0.53 & 0.48 & 0.50 & 0.65 & 0.65 & \textbf{0.65}  \\
\hline 
\end{tabular}
}
\caption{\label{tab:human}Precision, Recall and F-Measure of human evaluation and the character-based B-LSTM for the 5 most frequent emojis and 1,000 tweets.}
\end{table}

\subsection{Second Experiment}
\label{sec:human}
Given that \newcite{miller2016blissfully} pointed out that people tend to give multiple interpretations to emojis, we carried out an experiment in which we evaluated human and machine performances on the same task.
We randomly selected 1,000 tweets from our test set of the 5 most frequent emojis used in the previous experiment, and asked humans to  predict, after reading a tweet (with the emoji removed),  the emoji the text evoked. We opted for the 5 emojis task to reduce annotation efforts. After displaying the text of the tweet, we asked the human annotators ``What is the emoji you would include in the tweet?'', and gave the possibility to pick one of 5 possible emojis 
\includegraphics[height=0.32cm,width=0.32cm]{1f602.png},
\includegraphics[height=0.32cm,width=0.32cm]{2764.png},
\includegraphics[height=0.32cm,width=0.32cm]{1f60d.png},
\includegraphics[height=0.32cm,width=0.32cm]{1f4af.png}, and
\includegraphics[height=0.32cm,width=0.32cm]{1f525.png}.
Using the crowdsourcing platform `'CrowdFlower'', we designed an experiment where the same tweet was presented to four annotators (selecting the final label by majority agreement). Each annotator assessed a maximum of 200 tweets. The annotators were selected from the United States of America and of high quality (level 3 of CrowdFlower). One in every ten tweets, was an obvious test question, and annotations from subjects who missed more than 20\% of the test questions were discarded. The overall inter-annotator agreement was 73\% (in line with previous findings \cite{miller2016blissfully}). After creating the manually annotated dataset, we compared the human annotation and the char-BLSTM model with the gold standard (i.e. the emoji used in the tweet).

\begin{figure}
\centering
\includegraphics[height=4.1cm,keepaspectratio]{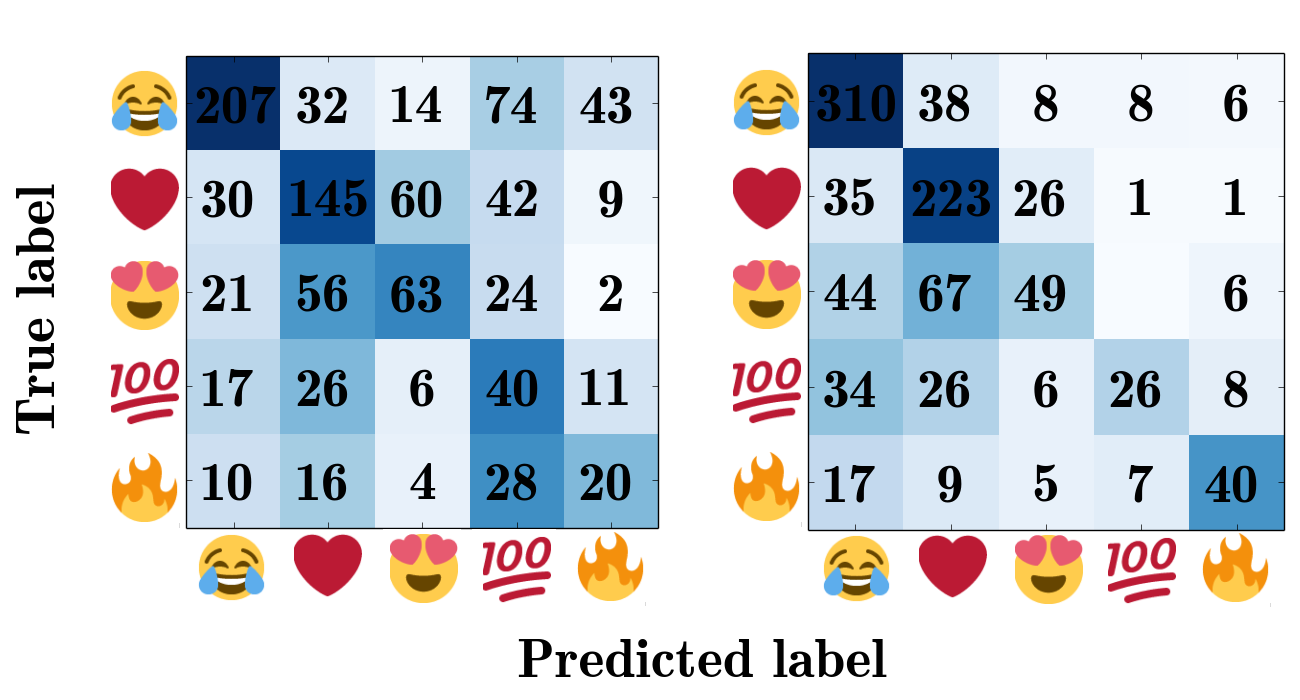} \\ 
\caption{\label{tab:cf}Confusion matrix of the second experiment. On the left the human evaluation and on the right the char-BLSTM model.}
\end{figure}

We can see in Table \ref{tab:human}, where the results of the comparison are presented, that the char-BLSTM performs better than humans, with a F1 of 0.65 versus 0.50. The emojis that the char-BLSTM struggle to predict are
\includegraphics[height=0.32cm,width=0.32cm]{1f60d.png}
and
\includegraphics[height=0.32cm,width=0.32cm]{1f4af.png}
, while the human annotators mispredict
\includegraphics[height=0.32cm,width=0.32cm]{1f4af.png}
and
\includegraphics[height=0.32cm,width=0.32cm]{1f525.png}
mostly. We can see in the confusion matrix of Figure \ref{tab:cf} that
\includegraphics[height=0.32cm,width=0.32cm]{1f60d.png}
is misclassified as 
\includegraphics[height=0.32cm,width=0.32cm]{2764.png} 
by both human and LSTM, and the 
\includegraphics[height=0.32cm,width=0.32cm]{1f4af.png}
emoji is mispredicted as 
\includegraphics[height=0.32cm,width=0.32cm]{1f602.png} and
\includegraphics[height=0.32cm,width=0.32cm]{2764.png}.
An interesting result is the number of times  
\includegraphics[height=0.32cm,width=0.32cm]{1f4af.png}
was chosen by human annotators; this emoji occurred 100 times (by chance) in the test set, but it was chosen 208 times, mostly when the correct label was the laughing emoji 
\includegraphics[height=0.32cm,width=0.32cm]{1f602.png}. 
We do not observe the same behavior in the char-BLSTMs, perhaps because they encoded information about the probability of these two emojis and when in doubt, the laughing emoji was chosen as more probable.


\section{Conclusions}
\label{sec:conclusions}
Emojis are used extensively in social media, however little is known about their use and semantics, especially because emojis are used differently over different communities \cite{barbieri2016revealing,barbieri2016cosmopolitan}. In this paper, we provide a neural architecture to model the semantics of emojis, exploring the relation between words and emojis. We proposed for the first time an automatic method to, given a tweet, predict the most probable emoji associated with it. We showed that the LSTMs outperform humans on the same emoji prediction task, suggesting that automatic systems are better at generalizing the usage of emojis than humans. Moreover, the good accuracy of the LSTMs suggests that there is an important and unique relation between sequences of words and emojis. 


As future work, we  plan to make the model able to predict more than one emoji per tweet, and explore the position of the emoji in the tweet, as close words can be an important clue for the emoji prediction task.

\section*{Acknowledgments}
We thank the three reviewers for their time and their useful suggestions. First and third authors acknowledge support from the TUNER project (TIN2015-65308-C5-5-R, MINECO/FEDER, UE) 
and the Maria de Maeztu Units of Excellence Programme (MDM-2015-0502).

\bibliography{emnlp2016}\bibliographystyle{eacl2017}

\end{document}